\documentclass[conference]{IEEEtran}

\usepackage{booktabs}
\usepackage{multirow}
\usepackage{graphicx}
\usepackage{xspace}
\usepackage{subfigure}
\usepackage{url}
\begin{document}

\def \eg {\emph{e.g.}, }
\def \ie {\emph{i.e.}, }
\def \et {\emph{et al.}}
\def \th {\mathop{\mathrm{th}}}

\newcommand{\tommy}[1]{{\color{orange}Tommy: #1}}
\newcommand{\shengzhong}[1]{{\color{blue}Shengzhong: #1}}
\newcommand{\maggie}[1]{{\color{green}Maggie: #1}}

\newcommand{\todo}[1]{{\color{red}\textbf{TODO: #1}}}
\newcommand{\note}[1]{{\color{purple}\textbf{CITE: #1}}}

\newcommand{\firstday}{Control Set\xspace}
\newcommand{\secondday}{Noisy Set\xspace}

\title{On the Efficiency and Robustness of Vibration-based Foundation Models for IoT Sensing: A Case Study}

\author{
    Tomoyoshi Kimura$^{\dagger}$,
    Jinyang Li$^{\dagger}$,
    Tianshi Wang$^{\dagger}$,
    Denizhan Kara$^{\dagger}$ 
    Yizhuo Chen$^{\dagger}$,
    Yigong Hu$^{\dagger}$,
    Ruijie Wang$^{\dagger}$, \\
    Maggie Wigness$^{*}$,
    Shengzhong Liu$^{\ddagger}$,
    Mani Srivastava$^{\S}$, 
    Suhas Diggavi$^{\S}$,
    Tarek Abdelzaher$^{\dagger}$\\
    
    \IEEEauthorblockA{$^{\dagger}$\textit{University of Illinois at Urbana-Champaign}, $^{*}$\textit{DEVCOM Army Research Laboratory}}
    \IEEEauthorblockA{$^{\ddagger}$\textit{Shanghai Jiao Tong University}, $^{\S}$\textit{University of California, Los Angeles}}

    \{tkimura4, jinyang7, tianshi3, kara4, yizhuoc, yigongh2, ruijiew2\}@illinois.edu, \\
    \{maggie.b.wigness\}.civ@army.mil,
    shengzhong@sjtu.edu.cn, \\
    \{mbs, suhas\}@ee.ucla.edu,
    zaher@illinois.edu
}

\maketitle
\begin{abstract}
This paper demonstrates the potential of vibration-based Foundation Models (FMs), pre-trained with {\em unlabeled sensing data\/}, to improve the robustness of run-time inference in (a class of) IoT applications. A case study is presented featuring a vehicle classification application using acoustic and seismic sensing. 
The work is motivated by the success of foundation models in the areas of natural language processing and computer vision, leading to generalizations of the FM concept to other domains as well, where significant amounts of unlabeled data exist that can be used for self-supervised pre-training. One such domain is {\em IoT applications\/}. Foundation models for selected sensing modalities in the IoT domain can be pre-trained in an environment-agnostic fashion using available {\em unlabeled\/} sensor data and then fine-tuned to the deployment at hand using a small amount of labeled data. The paper shows that the pre-training/fine-tuning approach improves the robustness of downstream inference and facilitates adaptation to different environmental conditions. More specifically, we present a case study in a real-world setting to evaluate a simple (vibration-based) FM-like model, called FOCAL, demonstrating its superior robustness and adaptation, compared to conventional supervised deep neural networks (DNNs).
We also demonstrate its superior convergence over supervised solutions.
Our findings highlight the advantages of vibration-based FMs (and FM-inspired self-supervised models in general) in terms of inference robustness, runtime efficiency, and model adaptation (via fine-tuning) in resource-limited IoT settings.
\end{abstract}


\section{Introduction}\label{sec:introduction}
The paper presents a real-world case study of a target classification application, based on seismic and acoustic sensing, that demonstrates how a self-supervised neural network model pre-trained with {\em unlabeled sensor data\/} (using pre-training techniques common to foundation models~\cite{liu2023focal}) can significantly improve run-time inference robustness and adaptation. Modalities, such as acoustic or seismic sensing, are particularly sensitive to environmental factors. Even in the same application domain, such as target tracking, a target (e.g., some vehicle on a road) may generate different sensory signatures depending on a variety of factors, such as the type of terrain (paved road, gravel, sand, ...), background noise (rain, wind, construction, traffic, ...), and other natural and/or human disturbances. Training an inference task (e.g., a target classifier) to handle all such contingencies is a daunting undertaking. Inspired by pre-training solutions used for foundation models, can one pre-train a general target-independent and environment-independent model once, based on large amounts of unlabeled data (henceforth called a {\em foundation model\/}), then fine-tune it in a very light-weight fashion to each deployment environment and set of targets of interest?

Early {\em supervised\/} solutions for intelligent IoT applications are label-hungry due to the large sizes of modern deep neural networks (DNNs) that call for commensurately large volumes of (labeled) input training data. In the absence of sufficient amounts of labeled data, supervised neural-network training techniques suffer from overfitting, thereby dramatically reducing the robustness of run-time inference~\cite{wang2022methodological}. In contrast, by obviating the need for labeled data in pre-training (and requiring only small amounts of labeled data for fine-tuning), foundation models developed for intelligent IoT applications can improve inference robustness and adaptation to domain shifts and environmental noise.

Unlike supervised training techniques that directly teach a neural network {\em how to perform a particular inference task\/}, 
a foundation model is the output of (pre-)training that aims to teach the neural network {\em a better internal representation of domain-specific data\/}. By empirically learning statistical properties and patterns found in large domain-specific datasets, such an internal representation encodes (empirical approximations of) higher-level semantics or ``knowledge" of the domain. Clearly, the degree to which such outcomes can be elicited depends on the amount of data used. Three important features thus characterize the pre-training of foundation models. First, it is {\em self-supervised}; no labeled data are needed. Second, it is {\em task-agnostic\/}; it does not know the downstream inference task(s) and, as such, can in principle support several different tasks, deployments, or environments. Finally, it generally uses a large amount of (unlabeled) data. For the sake of a proof of concept, we sacrifice the last property a bit in this study. 
The feasibility of pre-training in the absence of labeled data and without knowing the exact downstream task(s) makes the approach attractive to IoT applications, especially from a robustness perspective. 
Interestingly, despite the use of a smaller (and thus more manageable) amount of pre-training data in this paper, the robustness advantages of the resulting model are still possible to illustrate.
We show that the pre-trained model can be fine-tuned with only a minimal amount of labeled data for a specific downstream deployment,
allowing more robust classification than baseline (supervised) approaches. 


Another challenge for IoT sensing is the computational limitations in IoT devices.
Rapid advances in computational resources have led to increasingly large DNNs~\cite{brown2020language}. 
However, many IoT devices remain limited by their resource constraints~\cite{chatterjee2019context}.
These devices, from simple sensors to complex wearables, often lack the necessary processing power, memory, and energy efficiency to support the training and operation of large-scale DNNs in real-time.
This discrepancy poses significant challenges for deploying and training advanced DNNs in IoT applications \cite{yao2018fastdeepiot}, introducing bottlenecks to the model performance on IoT devices.
We show that the pre-trained model we use is capable of execution on a Raspberry-Pi class of devices in real-time with a higher fine-tuning convergence rate while offering more robust performance than its supervised counterparts.




The rest of the paper is organized as follows. We cover a brief background on foundation model pre-training and the specific model used in this paper in Section~\ref{sec:background}. 
We describe our case study and experimental set-up in Section~\ref{sec:setup}. Section~\ref{sec:evaluation} presents the evaluation results, followed by discussion in Section~\ref{sec:discussion}. Section~\ref{sec:related} covers related work.  Section~\ref{sec:conclusion} concludes the paper.
\vspace{-0.1in}
\section{Self-Supervised Model Pre-Training}
\label{sec:background}
While many techniques were proposed recently for self-supervised pre-training of foundation models, two are particularly widespread: learning to reconstruct masked~\cite{devlin2018bert,he2022masked} (or distorted~\cite{lewis2019bart}) inputs and contrastive learning~\cite{chen2020simple,you2020graph,chuang2020debiased, liu2021contrastive}. They differ in the way they train the model useful concepts from the domain, without the need for labeled data. Specifically, masking/distortion removes/distorts parts of the input, and then rewards the model for the correct reconstruction of these parts. Clearly, a model that learns correct reconstruction must have encapsulated some knowledge about the target domain. Contrastive learning teaches the model what ``similarity" means in the target domain (by contrasting similar and dissimilar sample pairs), such that similar inputs are grouped closer together in a latent space. To do so without labels, it often relies on semantics-invariant input transformations that convert individual input samples to ``similar" ones (without necessarily knowing what the sample labels or semantics are). An example of such transformations in vision is image resizing. An example in time-series data is adding simulated noise. The result of rewarding the model for putting similar samples closer together in the latent space is a well-organized learned latent representation, where proximity implies semantic similarity.

In this paper, for pre-training, we use a contrastive learning framework, called FOCAL \cite{liu2023focal}, recently proposed for (pre-training in) intelligent multimodal sensing applications.
FOCAL pre-trains {\em an encoder\/} to extract a structured latent representation of the input multimodal sensing data. This latent representation separates shared and private subspaces.
The shared subspace contains common information shared across the different sensing modalities. The private subspaces hold additional modality-exclusive information. 
An orthogonality constraint is applied among the private subspaces, as well as between each private subspace and the shared subspace to enforce information independence among these subspaces.
A pre-trained encoder is fine-tuned by appending a single linear layer whose weights are adapted to the downstream use scenario (using a small amount of task-specific labeled data).


We utilize FOCAL to train two popular DNN encoders (DeepSense \cite{yao2017deepsense} and SWIN-Transformer~\cite{liu2021swin}) on a multi-modal Moving Object Detection~\cite{liu2023focal} (MOD) dataset that consists of acoustic and seismic signals. Then, we perform a two-day experiment in a real-world neighborhood as a case study to examine the performance of FOCAL against supervised counterparts. The experimental setting and results are described below.

\section{An Experimental Study}\label{sec:setup}

\begin{table*}[!t]
\centering
\caption{
Training Configurations: Below, we detail the training parameters including the batch size (number of samples per batch), the optimizer for updating model parameters, the initial learning rate (LR), and the LR scheduler for dynamic LR adjustments, alongside its LR decay rate. The table also lists the total training epochs and the data augmentations applied.
}
\label{tab:training_config}
\resizebox{\textwidth}{!}{%
\begin{tabular}{@{}c|c|c|c|c|c|c|c@{}}
\toprule
Stage      & Batch Size & Optimizer & Initial LR  & LR Scheduler & LR Decay & Epochs & Augmentations      \\ \midrule
Supervised & 128        & AdamW \cite{loshchilov2018decoupled} & 1e-4 & Cosine \cite{loshchilov2016sgdr}    & 0.2    & 500          & Mixup, Phase Shift \\
Pretrain &
  256 &
  AdamW \cite{loshchilov2018decoupled} &
    0.0001 &
  Cosine \cite{loshchilov2016sgdr} &
  0.05 &
  6000 &
  \begin{tabular}[c]{@{}c@{}}Permutation, Negation, Time Warp, Horizontal Flip,\\ Magnitude Warp, Scaling,  Phase Shift\end{tabular} \\
fine-tune   & 256        & Adam \cite{KingBa15}     & 1e-3 & Cosine \cite{loshchilov2016sgdr}    & 0.2    & 200          & Mixup, Phase Shift \\ \bottomrule
\end{tabular}%
}
\end{table*}
Our experiment was conducted at an outdoor research facility located on (repurposed) state park grounds. Sensors were deployed and vehicles were driven past the sensors over a period spanning two days. On the first day, the environment surrounding the experiment was controlled and disturbance-free. The second day featured significant interference (as described in the next section). 

FOCAL was {\em pre-trained\/} 
on a previously published dataset~\cite{liu2023focal} collected from acoustic and seismic sensors, deployed in different urban and rural environments that varied in terrain (paved, gravel, dirt, rooftop, etc) and environmental conditions (quiet, windy, etc), recording the passage of a variety of target types, mostly focusing on civilian automobiles, bikes, and humans. The pre-training data did {\em not\/} include any from the deployment reported in this paper. To experiment with the robustness of the pre-trained model, we {\em fine-tune\/} it on part of the data collected in the new deployment and {\em test\/} the fine-tuned model's performance under the same or different deployment conditions. A comparison is carried out with supervised approaches.

We show that FOCAL is more robust to domain changes than other methods.
Additionally, we demonstrate FOCAL's superiority in terms of label efficiency by comparing performance under different amounts of labels used for fine-tuning.
Finally, we present our findings on the computational efficiency of FOCAL and explore the potential applications for real-time IoT systems.

\subsection{The Experimental Setup}
\begin{figure}[t!]
  \centering
  \frame{\includegraphics[width=0.80\linewidth]{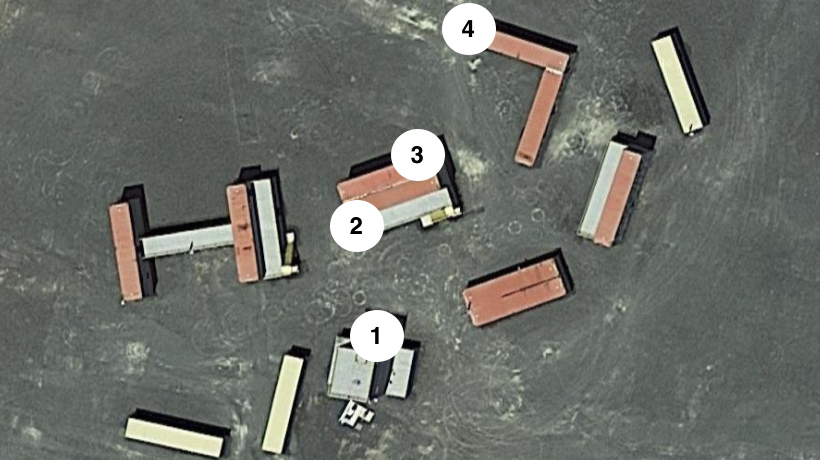}}
  \caption{The satellite view of the case study neighborhood with labeled nodes. }
  \label{fig:satellite_view}
\end{figure}

The experiment was performed using four deployed multimodal sensor nodes.
Figure~\ref{fig:satellite_view} shows a satellite view of the facility and the four locations where we set up the sensor nodes. Nodes 1 \& 4 utilized the RaspberryShake\footnote{\url{https://raspberryshake.org/}} 4D, while Nodes 2 \& 3 utilized the RaspberryShake 1D.
Each node featured a geophone and a microphone array, collecting seismic and acoustic vibration signals from nearby objects. 
In each run, a specific target navigated the neighborhood, passing the sensors in some arbitrary order within a short time window. 
Four distinct target types were used: (i) a Polaris\footnote{\url{https://www.polaris.com/}} off-road vehicle, (ii) a Warthog\footnote{\url{https://clearpathrobotics.com/warthog-unmanned-ground-vehicle-robot/}} all-terrain unmanned ground robot, (iii) a Husky unmanned outdoor field robot\footnote{\url{https://clearpathrobotics.com/husky-unmanned-ground-vehicle-robot/}}, and (iv) a standard civilian automobile.
We collected 23 runs in total, each lasting approximately 10 minutes.

Although the sensors and targets were identical on both days, the sensor data distributions of the two days varied substantially due to different on-site events.
On the first day, we conducted a controlled test run with only our operators around. 
On the second day, 5-6 research groups simultaneously worked on multiple experiments, which led to increased interference. Individuals walked and talked near our sensors, introducing human-related (acoustic and seismic) noise. Loud motor-powered generators were used by some teams creating additional acoustic and seismic interference. Strong wind on the second day further added to environmental disturbances. 
Thus, we partition the data collected by day. We refer to data collected on the first day as the \textit{\textbf{\firstday}} and data collected on the second day as the \textit{\textbf{\secondday}}. 

\subsection{Datasets}

We also consider the MOD dataset released in \cite{liu2023focal}. This dataset contains multi-modal seismic and acoustic signals describing nearby moving objects. 
Therefore, we have three sets of data subject to varying distribution shifts --- \textbf{\textit{MOD, \firstday, and \secondday}}.
We follow the same setup as \cite{liu2023focal} to process these three sets with a 0.2-second overlapping ratio between 2 seconds samples of 8000Hz acoustic 100Hz seismic data. 
We then partition MOD into a set of unlabeled data used to pre-train the FM and a set of labeled data for supervised training and fine-tuning. 
The distribution of the MOD dataset is significantly different compared to the other two sets due to distinct locations, sensor placements, and moving targets.
The \firstday and the \secondday, though have similar targets, have different distributions as well due to runtime conditions. 

\subsection{Training Pipelines}

\subsubsection{Training Frameworks}

We choose FOCAL~\cite{liu2023focal} as our self-supervised training framework.
We use FOCAL with two different backbone encoders:
\begin{itemize}
    \item \textbf{DeepSense} \cite{yao2017deepsense} is a DNN classifier designed for time-series sensory inputs. It applies convolution layers on modality spectrograms to extract general features and then utilizes recurrent layers (stacked GRU) to further extract global temporal relationships.
    \item \textbf{SWIN-Transformer (SW-T)} \cite{liu2021swin} is a variant of Vision Transformer (ViT)\cite{dosovitskiy2020image}, proposing to extract a hierarchical representation through downsampling and shifting window operations. Similar to ViT, it partitions the sample into patches. What makes SW-T different from ViT is that SW-T groups different patches into non-overlapping windows and computes self-attention within each window to minimize computation costs. These windows are further shifted to take advantage of the cross-window connection.
\end{itemize}


\subsubsection{Pretraining}
We pretrain FOCAL with the unlabeled set from the MOD dataset.
We randomly apply different time and frequency augmentations in the time domain and frequency domain. We use STFT to convert each modality sample into the frequency domain and then extract the modality embedding. Training configurations used during pre-training are presented in Table \ref{tab:training_config}.

\subsubsection{Supervised Training/fine-tuning}
In the fine-tuning stage, we use labeled samples to perform supervised fine-tuning on the pretrained model. We freeze the pretrained model and add a linear layer for target classification (from the concatenated modality embeddings). We would like to note that only the linear layer is trained at the fine-tuning stage. 
During fine-tuning, We apply mixup \cite{zhang2018mixup} augmentation in the time domain and phase shift augmentation in the frequency domain. 
We also separately train supervised DNNs for the two backbone encoders as the benchmarks. The supervised model contains an additional fusion layer to fuse the modality embeddings for classification. 
Training configurations for fine-tuning and supervised benchmark can be found in Table \ref{tab:training_config}. We also use a
supervised model initially trained on the MOD dataset and later
fine-tuned on its final classification layer, mirroring FOCAL’s fine-tuning approach. We call it the supervised-fine-tuned baseline.

\begin{table*}[htb]
\centering
\caption{fine-tune Results on the \firstday}
\label{tab:retraining_day1_day1}
\resizebox{0.75\textwidth}{!}{%
\begin{tabular}{@{}cc|cc|cc|cc|cc@{}}
\toprule
\multicolumn{2}{c|}{Label Ratio} &
  \multicolumn{2}{c|}{100\%} &
  \multicolumn{2}{c|}{50\%} &
  \multicolumn{2}{c|}{10\%} &
  \multicolumn{2}{c}{1\%} \\ \midrule
\multicolumn{1}{c|}{Encoder} & Framework           & Acc    & F1     & Acc    & F1     & Acc             & F1              & Acc             & F1              \\ \midrule
\multicolumn{1}{c|}{\multirow{3}{*}{DeepSense}} &
  Supervised &
  \textbf{0.9684} &
  \textbf{0.9637} &
  \textbf{0.9425} &
  \textbf{0.9328} &
  0.8078 &
  0.7714 &
  0.5247 &
  0.5019 \\
\multicolumn{1}{c|}{}        & Supervised-fine-tune & 0.7933 & 0.7578 & 0.7762 & 0.7379 & 0.7383          & 0.6892          & 0.5974          & 0.5392          \\
\multicolumn{1}{c|}{}        & FOCAL               & 0.9330 & 0.9293 & 0.9204 & 0.9154 & \textbf{0.8976} & \textbf{0.8893} & \textbf{0.8078} & \textbf{0.7876} \\ \midrule
\multicolumn{1}{c|}{\multirow{3}{*}{SW-T}} &
  Supervised &
  \textbf{0.9842} &
  \textbf{0.9840} &
  \textbf{0.9608} &
  \textbf{0.9589} &
  0.7434 &
  0.7107 &
  0.3660 &
  0.2802 \\
\multicolumn{1}{c|}{}        & Supervised-fine-tune & 0.6372 & 0.5829 & 0.6327 & 0.5778 & 0.6056          & 0.5592          & 0.5607          & 0.5037          \\
\multicolumn{1}{c|}{}        & FOCAL               & 0.9526 & 0.9473 & 0.9558 & 0.9524 & \textbf{0.9425} & \textbf{0.9372} & \textbf{0.8312} & \textbf{0.8176} \\ \bottomrule
\end{tabular}%
}
\end{table*}

\begin{table*}[htb]
\centering
\caption{Test Results on the \secondday}
\label{tab:retraining_day1_day2}
\resizebox{0.75\textwidth}{!}{%
\begin{tabular}{@{}cc|cc|cc|cc|cc@{}}
\toprule
\multicolumn{2}{c|}{Label Ratio} &
  \multicolumn{2}{c|}{100\%} &
  \multicolumn{2}{c|}{50\%} &
  \multicolumn{2}{c|}{10\%} &
  \multicolumn{2}{c}{1\%} \\ \midrule
\multicolumn{1}{c|}{Encoder} &
  Framework &
  Acc &
  F1 &
  Acc &
  F1 &
  Acc &
  F1 &
  Acc &
  F1 \\ \midrule
\multicolumn{1}{c|}{\multirow{3}{*}{DeepSense}} &
  Supervised &
  \textbf{0.6769} &
  \textbf{0.6843} &
  \textbf{0.6603} &
  \textbf{0.6639} &
  0.5805 &
  0.5764 &
  0.4688 &
  0.4919 \\
\multicolumn{1}{c|}{} &
  Supervised-fine-tune &
  0.5766 &
  0.5735 &
  0.5689 &
  0.5650 &
  0.5539 &
  0.5458 &
  0.4358 &
  0.4041 \\
\multicolumn{1}{c|}{} &
  FOCAL &
  0.6558 &
  0.6640 &
  0.6515 &
  0.6601 &
  \textbf{0.6578} &
  \textbf{0.6634} &
  \textbf{0.6101} &
  \textbf{0.6153} \\ \midrule
\multicolumn{1}{c|}{\multirow{3}{*}{SW-T}} &
  Supervised &
  0.5454 &
  0.5397 &
  0.5126 &
  0.5040 &
  0.4180 &
  0.3962 &
  0.2838 &
  0.2157 \\
\multicolumn{1}{c|}{} &
  Supervised-fine-tune &
  0.4179 &
  0.3968 &
  0.4149 &
  0.3944 &
  0.4072 &
  0.3883 &
  0.3862 &
  0.3527 \\
\multicolumn{1}{c|}{} &
  FOCAL &
  \textbf{0.6641} &
  \textbf{0.6788} &
  \textbf{0.6742} &
  \textbf{0.6819} &
  \textbf{0.6924} &
  \textbf{0.7050} &
  \textbf{0.5549} &
  \textbf{0.5508} \\ \bottomrule
\end{tabular}%
}
\end{table*}

\section{Evaluation Results}\label{sec:evaluation}
Below, we examine FOCAL performance after fine-tuning with some target domain data then compare the computational efficiency of the supervised and the Foundation models. 



\subsection{Model Retraining/fine-tuning}



We divide the \firstday into training, validation, and testing data with a ratio of 8:1:1.
We train and fine-tune the models using different amounts of labeled samples from the training data of the \firstday (100\%, 50\%, 10\%, 1\%) and then evaluate their performance on the testing data from the same set.
Table \ref{tab:retraining_day1_day1} summarizes the performance of the retrained DNNs on the \firstday, under different label ratios. When the amount of labeled data used is high (100\% or 50\%), the supervised approaches work well. In fact, they slightly outperform FOCAL (that tunes its last layer only). However, as the amount of labeled data decreases (10\% and 1\%), the supervised approaches degrade substantially, whereas FOCAL suffers a much lower penalty in performance, suggesting a higher label efficiency. Note also that the supervised-fine-tuned benchmark is dominated by FOCAL, offering no advantage across the board.
The gap between the supervised-fine-tuned benchmark and FOCAL underscores FM's ability to encode prior knowledge much more robustly than the supervised models. As such, it can be more easily adapted to various IoT deployment conditions with a minimal amount of labels. 

Next, we use the models trained on the \firstday and evaluate them on the \secondday.
While the \firstday and the \secondday contain identical target objects collected on the same set of sensors, the dynamic nature of the IoT System can still affect the collected data distribution. 
We examine such domain shift effect in Table \ref{tab:retraining_day1_day2}. 
As we lower the label ratio, supervised models experience significant degradation, whereas FOCAL remains relatively stable. 
As before, FOCAL dominates the supervised-fine-tuned approach, suggesting that its self-supervised pre-training has better knowledge transfer.


\subsection{Training Efficiency}

\begin{figure}[t!]
  \centering
  \subfigure[DeepSense]{\includegraphics[width=0.49\columnwidth]{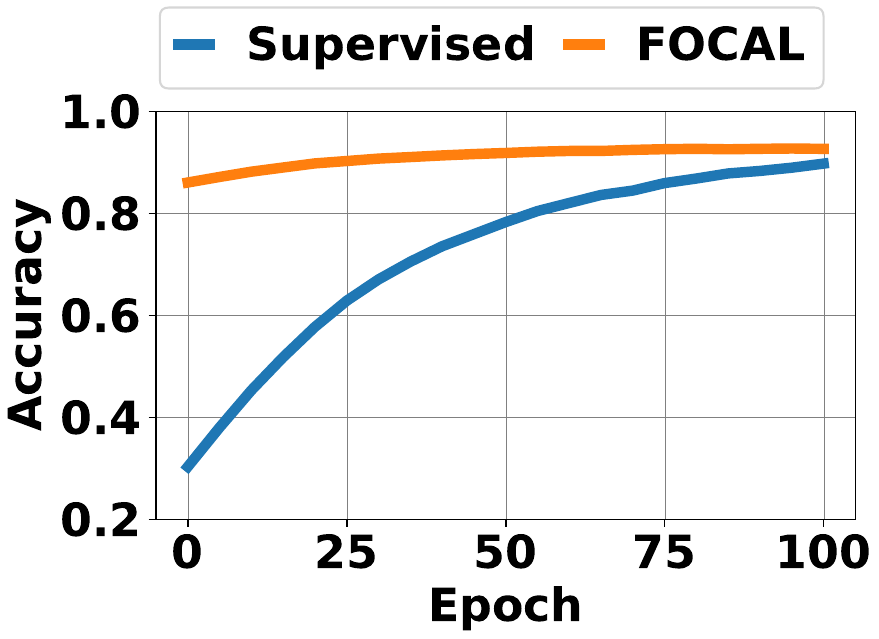}}
  \subfigure[SW-T]{\includegraphics[width=0.49\columnwidth]{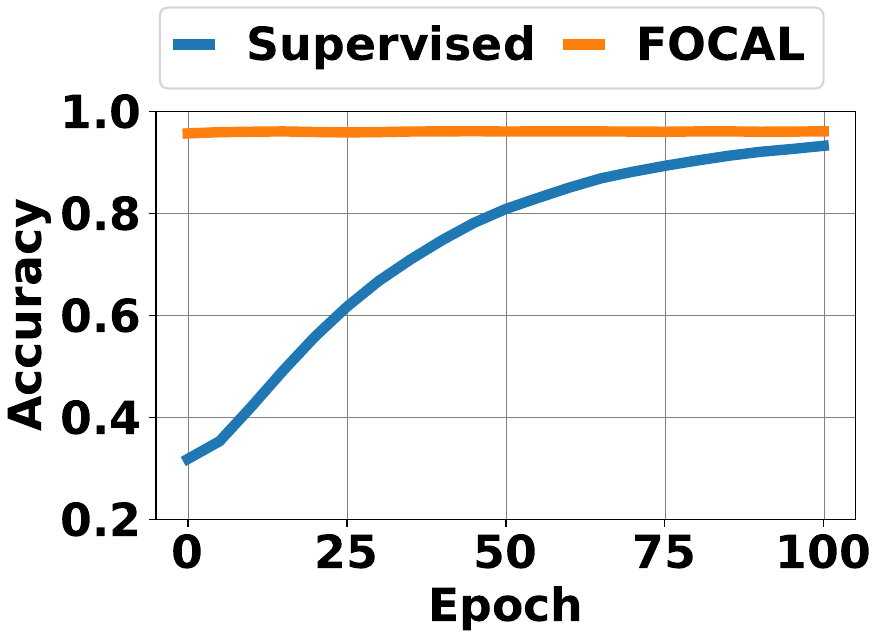}}
  \caption{Accuracy curves of Supervised Training and FOCAL fine-tuning.} 
  \label{fig:convergence}
\end{figure}

In this section, we compare the training efficiency of the supervised models and the fine-tuning efficiency of FOCAL (which we refer to as ``training" efficiency as well, for the sake of brevity, below). 
We define the training efficiency as the convergence speed or the number of training epochs needed for convergence.
As shown in Table~\ref{tab:retraining_day1_day1}, both the supervised model and FOCAL perform well after training on the \firstday. 
We compare their convergence speed by observing the training accuracy curves in Figure~\ref{fig:convergence} during the first 100 epochs. 
On both backbone encoders, FOCAL (fine-tuning) converges much faster with near-optimal performance achieved in the first few epochs, compared to training the supervised model.
This shows that the pre-trained representation is useful for the downstream task and can easily transfer knowledge to achieve high performance in a short time.
On the other hand, since the supervised models are trained from scratch, they begin at a lower accuracy and with more parameters to train (FOCAL only updates the linear classification layer during fine-tuning, as opposed to the supervised benchmark that trains all its parameters). Thus, the supervised algorithm approaches FOCAL performance only towards the end of the 100 epochs. 
We do not consider the supervised-fine-tune benchmarks since they are dominated by the others. 
While not shown, FOCAL also requires much less memory to fine-tune its single last layer, compared to retraining an entire supervised model from scratch.

\section{Discussion}\label{sec:discussion}

The experimental study reported in this paper suggests that the task-agnostic nature of pre-training of self-supervised models endows them with greater robustness, making them ideally suited for IoT application deployment across various environments with only limited fine-tuning needed to achieve high-quality inference.
Unlike traditional supervised models, these pre-trained models exploit large amounts of unlabeled data, leading to enhanced resilience against domain shifts.
Although this paper only leverages a small-scale unlabeled dataset for pre-training, the pre-trained models already exhibit characteristics of foundation models with great robustness in different domains. 
This is particularly useful in IoT sensing scenarios where different sensor deployments (even within the same application) may be subjected to vastly different conditions.
For example, target tracking using acoustic/vibration sensing will see significant distributional shifts across urban areas, rural roads, freeways, gravel parking lots, etc, as well as across different weather conditions (wind, snow, rain), and different target types. 
Pre-training the foundation model with a larger scale dataset and larger backbone encoders could also potentially improve its downstream robustness, and we leave that to future work.



The high label efficiency of pre-trained models further facilitates their rapid deployment to a wide array of downstream tasks, where label scarcity is a critical challenge.
These models exhibit exceptional adaptability to varying physical environments, which makes them suitable to meet the demands of real-time CPS. Merely training a single linear layer in FOCAL, for fine-tuning, can easily reach optimal performance within a few epochs. This efficiency not only enhances the practicality of FMs in dynamic settings but also opens opportunities for on-device training, making it feasible to train FMs on resource-constrained IoT devices.

\section{Related Work}\label{sec:related}
 Deep Learning has catalyzed significant advances in inference from IoT sensing data~\cite{book},  with DNNs becoming integral to a wide range of IoT applications \cite{salehi2022deep, radu2018multimodal}. 
However, domain-specific challenges still lead to many limitations in building robust DNNs for IoT sensing.
Deployed DNNs must handle unpredictable interference in the field that greatly alters the statistical distribution of collected sensor data. 
The altered distribution, or \textit{domain shift}, can significantly degrade DNN performance, leading to inaccurate results and potentially severe consequences.

More recently, Foundation Models (FMs) \cite{bommasani2021opportunities} have gained increasing popularity, most notably in language \cite{devlin2018bert, radford2019language} and vision \cite{he2022masked, oquab2023dinov2}.
The techniques were then generalized to other areas where domain-specific FMs emerged, such as security~\cite{almaraz2023enhancing, zhang2024market}, networking~\cite{zhang2023self, towhid2022encrypted}, and meteorology~\cite{mai2023opportunities}.


Contrastive Learning (CL) \cite{liu2023focal, chen2021mocov3, ouyang2022cosmo, poklukar2022gmc, tonekaboni2021unsupervised, yue2022ts2vec} has been a popular form of SSL to extract a robust embedding space during pre-training.
The main idea is to pull similar samples closer while pushing other samples further apart in the embedding space.
Unimodal CL frameworks like \cite{chen2020simclr, chen2021mocov3} apply random augmentations to learn transformation invariant information.
\cite{tian2020contrastive, poklukar2022gmc, ouyang2022cosmo} are multi-modal CL frameworks that enforce cross-modal consistency. CL for time series has also been extensively studied in \cite{tonekaboni2021unsupervised, yue2022ts2vec}.


Improving resilience against domain shifts has been widely studied in recent years. 
\cite{li2021faster, zhao2022memory} investigate improving the efficiency of unsupervised domain adaptation for IoT applications. However, these works primarily consider classifiers trained in a supervised manner. 
Others have also worked on Federated Learning-based domain generalization \cite{zhang2023federated, huang2022incremental}.
Numerous works analyze SSL for domain generalization \cite{achituve2021self, xu2019self}, but less has been explored for IoT applications.


\section{Conclusions}\label{sec:conclusion}
In this paper, we examined an FM-based approach, specifically FOCAL, against conventional supervised models in the context of IoT sensing. Through our real-world case study, we have demonstrated how Foundation Models require minimal domain-specific tuning while allowing robust real-time inference. Our results highlight promising opportunities for Foundation Models in the IoT landscape. Our future work will focus on developing more scalable Foundation Models for generalized IoT systems. 
\section{Acknowledgements}
Research reported in this paper was sponsored in part by DEVCOM ARL under Cooperative Agreement W911NF-17-2-0196 (ARL IoBT CRA).
The views and conclusions contained in this document are those of the authors and should not be interpreted as representing the official policies, either expressed or implied, of the Army Research Laboratory or the U.S.Government.
The U.S. Government is authorized to reproduce and distribute reprints for Government purposes notwithstanding any copyright notation herein. 

\bibliographystyle{IEEEtran}
\bibliography{main.bib}

\begin{thebibliography}{10}
\providecommand{\url}[1]{#1}
\csname url@samestyle\endcsname
\providecommand{\newblock}{\relax}
\providecommand{\bibinfo}[2]{#2}
\providecommand{\BIBentrySTDinterwordspacing}{\spaceskip=0pt\relax}
\providecommand{\BIBentryALTinterwordstretchfactor}{4}
\providecommand{\BIBentryALTinterwordspacing}{\spaceskip=\fontdimen2\font plus
\BIBentryALTinterwordstretchfactor\fontdimen3\font minus \fontdimen4\font\relax}
\providecommand{\BIBforeignlanguage}[2]{{%
\expandafter\ifx\csname l@#1\endcsname\relax
\typeout{** WARNING: IEEEtran.bst: No hyphenation pattern has been}%
\typeout{** loaded for the language `#1'. Using the pattern for}%
\typeout{** the default language instead.}%
\else
\language=\csname l@#1\endcsname
\fi
#2}}
\providecommand{\BIBdecl}{\relax}
\BIBdecl

\bibitem{liu2023focal}
S.~Liu, T.~Kimura, D.~Liu, R.~Wang, J.~Li, S.~Diggavi, M.~Srivastava, and T.~Abdelzaher, ``Focal: Contrastive learning for multimodal time-series sensing signals in factorized orthogonal latent space,'' in \emph{Advances in Neural Information Processing Systems}, 2023.

\bibitem{wang2022methodological}
T.~Wang, D.~Kara, J.~Li, S.~Liu, T.~Abdelzaher, and B.~Jalaian, ``The methodological pitfall of dataset-driven research on deep learning: An iot example,'' in \emph{MILCOM 2022-2022 IEEE Military Communications Conference (MILCOM)}.\hskip 1em plus 0.5em minus 0.4em\relax IEEE, 2022, pp. 1082--1087.

\bibitem{brown2020language}
T.~Brown, B.~Mann, N.~Ryder, M.~Subbiah, J.~D. Kaplan, P.~Dhariwal, A.~Neelakantan, P.~Shyam, G.~Sastry, A.~Askell \emph{et~al.}, ``Language models are few-shot learners,'' \emph{Advances in neural information processing systems}, vol.~33, pp. 1877--1901, 2020.

\bibitem{chatterjee2019context}
B.~Chatterjee, N.~Cao, A.~Raychowdhury, and S.~Sen, ``Context-aware intelligence in resource-constrained iot nodes: Opportunities and challenges,'' \emph{IEEE Design \& Test}, vol.~36, no.~2, pp. 7--40, 2019.

\bibitem{yao2018fastdeepiot}
S.~Yao, Y.~Zhao, H.~Shao, S.~Liu, D.~Liu, L.~Su, and T.~Abdelzaher, ``Fastdeepiot: Towards understanding and optimizing neural network execution time on mobile and embedded devices,'' in \emph{Proceedings of the 16th ACM Conference on Embedded Networked Sensor Systems}, 2018, pp. 278--291.

\bibitem{devlin2018bert}
J.~Devlin, M.-W. Chang, K.~Lee, and K.~Toutanova, ``Bert: Pre-training of deep bidirectional transformers for language understanding,'' \emph{arXiv preprint arXiv:1810.04805}, 2018.

\bibitem{he2022masked}
K.~He, X.~Chen, S.~Xie, Y.~Li, P.~Doll{\'a}r, and R.~Girshick, ``Masked autoencoders are scalable vision learners,'' in \emph{Proceedings of the IEEE/CVF conference on computer vision and pattern recognition}, 2022, pp. 16\,000--16\,009.

\bibitem{lewis2019bart}
M.~Lewis, Y.~Liu, N.~Goyal, M.~Ghazvininejad, A.~Mohamed, O.~Levy, V.~Stoyanov, and L.~Zettlemoyer, ``Bart: Denoising sequence-to-sequence pre-training for natural language generation, translation, and comprehension,'' \emph{arXiv preprint arXiv:1910.13461}, 2019.

\bibitem{chen2020simple}
T.~Chen, S.~Kornblith, M.~Norouzi, and G.~Hinton, ``A simple framework for contrastive learning of visual representations,'' in \emph{International conference on machine learning}.\hskip 1em plus 0.5em minus 0.4em\relax PMLR, 2020, pp. 1597--1607.

\bibitem{you2020graph}
Y.~You, T.~Chen, Y.~Sui, T.~Chen, Z.~Wang, and Y.~Shen, ``Graph contrastive learning with augmentations,'' \emph{Advances in neural information processing systems}, vol.~33, pp. 5812--5823, 2020.

\bibitem{chuang2020debiased}
C.-Y. Chuang, J.~Robinson, Y.-C. Lin, A.~Torralba, and S.~Jegelka, ``Debiased contrastive learning,'' \emph{Advances in neural information processing systems}, vol.~33, pp. 8765--8775, 2020.

\bibitem{liu2021contrastive}
D.~Liu, T.~Wang, S.~Liu, R.~Wang, S.~Yao, and T.~Abdelzaher, ``Contrastive self-supervised representation learning for sensing signals from the time-frequency perspective,'' in \emph{2021 International Conference on Computer Communications and Networks (ICCCN)}.\hskip 1em plus 0.5em minus 0.4em\relax IEEE, 2021, pp. 1--10.

\bibitem{yao2017deepsense}
S.~Yao, S.~Hu, Y.~Zhao, A.~Zhang, and T.~Abdelzaher, ``Deepsense: A unified deep learning framework for time-series mobile sensing data processing,'' in \emph{International Conference on World Wide Web (WWW)}, 2017.

\bibitem{liu2021swin}
Z.~Liu, Y.~Lin, Y.~Cao, H.~Hu, Y.~Wei, Z.~Zhang, S.~Lin, and B.~Guo, ``Swin transformer: Hierarchical vision transformer using shifted windows,'' in \emph{IEEE/CVF International Conference on Computer Vision (CVPR)}, 2021.

\bibitem{loshchilov2018decoupled}
I.~Loshchilov and F.~Hutter, ``Decoupled weight decay regularization,'' in \emph{International Conference on Learning Representations}, 2018.

\bibitem{loshchilov2016sgdr}
------, ``Sgdr: Stochastic gradient descent with warm restarts,'' in \emph{International Conference on Learning Representations}, 2016.

\bibitem{KingBa15}
D.~Kingma and J.~Ba, ``Adam: A method for stochastic optimization,'' in \emph{International Conference on Learning Representations (ICLR)}, San Diega, CA, USA, 2015.

\bibitem{dosovitskiy2020image}
A.~Dosovitskiy, L.~Beyer, A.~Kolesnikov, D.~Weissenborn, X.~Zhai, T.~Unterthiner, M.~Dehghani, M.~Minderer, G.~Heigold, S.~Gelly \emph{et~al.}, ``An image is worth 16x16 words: Transformers for image recognition at scale,'' in \emph{International Conference on Learning Representations}, 2020.

\bibitem{zhang2018mixup}
H.~Zhang, M.~Cisse, Y.~N. Dauphin, and D.~Lopez-Paz, ``mixup: Beyond empirical risk minimization,'' in \emph{International Conference on Learning Representations}, 2018.

\bibitem{book}
M.~Srivatsa, T.~Abdelzaher, and T.~He, Eds., \emph{Artificial Intelligence for Edge Computing}.\hskip 1em plus 0.5em minus 0.4em\relax Springer, 2023.

\bibitem{salehi2022deep}
B.~Salehi, G.~Reus-Muns, D.~Roy, Z.~Wang, T.~Jian, J.~Dy, S.~Ioannidis, and K.~Chowdhury, ``Deep learning on multimodal sensor data at the wireless edge for vehicular network,'' \emph{IEEE Transactions on Vehicular Technology}, vol.~71, no.~7, pp. 7639--7655, 2022.

\bibitem{radu2018multimodal}
V.~Radu, C.~Tong, S.~Bhattacharya, N.~D. Lane, C.~Mascolo, M.~K. Marina, and F.~Kawsar, ``Multimodal deep learning for activity and context recognition,'' \emph{Proceedings of the ACM on interactive, mobile, wearable and ubiquitous technologies}, vol.~1, no.~4, pp. 1--27, 2018.

\bibitem{bommasani2021opportunities}
R.~Bommasani, D.~A. Hudson, E.~Adeli, R.~Altman, S.~Arora, S.~von Arx, M.~S. Bernstein, J.~Bohg, A.~Bosselut, E.~Brunskill \emph{et~al.}, ``On the opportunities and risks of foundation models,'' \emph{arXiv preprint arXiv:2108.07258}, 2021.

\bibitem{radford2019language}
A.~Radford, J.~Wu, R.~Child, D.~Luan, D.~Amodei, and I.~Sutskever, ``Language models are unsupervised multitask learners,'' 2019.

\bibitem{oquab2023dinov2}
M.~Oquab, T.~Darcet, T.~Moutakanni, H.~V. Vo, M.~Szafraniec, V.~Khalidov, P.~Fernandez, D.~Haziza, F.~Massa, A.~El-Nouby, R.~Howes, P.-Y. Huang, H.~Xu, V.~Sharma, S.-W. Li, W.~Galuba, M.~Rabbat, M.~Assran, N.~Ballas, G.~Synnaeve, I.~Misra, H.~Jegou, J.~Mairal, P.~Labatut, A.~Joulin, and P.~Bojanowski, ``Dinov2: Learning robust visual features without supervision,'' 2023.

\bibitem{almaraz2023enhancing}
J.~G. Almaraz-Rivera, J.~A. Cantoral-Ceballos, and J.~F. Botero, ``Enhancing iot network security: Unveiling the power of self-supervised learning against ddos attacks,'' \emph{Sensors}, vol.~23, no.~21, p. 8701, 2023.

\bibitem{zhang2024market}
Z.~Zhang, S.~Bu, Y.~Zhang, and Z.~Han, ``Market-level integrated detection against cyber attacks in real-time market operations by self-supervised learning,'' \emph{IEEE Transactions on Smart Grid}, 2024.

\bibitem{zhang2023self}
S.~Zhang, O.~T. Ajayi, and Y.~Cheng, ``A self-supervised learning approach for accelerating wireless network optimization,'' \emph{IEEE Transactions on Vehicular Technology}, 2023.

\bibitem{towhid2022encrypted}
M.~S. Towhid and N.~Shahriar, ``Encrypted network traffic classification using self-supervised learning,'' in \emph{2022 IEEE 8th International Conference on Network Softwarization (NetSoft)}.\hskip 1em plus 0.5em minus 0.4em\relax IEEE, 2022, pp. 366--374.

\bibitem{mai2023opportunities}
G.~Mai, W.~Huang, J.~Sun, S.~Song, D.~Mishra, N.~Liu, S.~Gao, T.~Liu, G.~Cong, Y.~Hu \emph{et~al.}, ``On the opportunities and challenges of foundation models for geospatial artificial intelligence,'' \emph{arXiv preprint arXiv:2304.06798}, 2023.

\bibitem{chen2021mocov3}
X.~Chen, S.~Xie, and K.~He, ``An empirical study of training self-supervised vision transformers,'' in \emph{IEEE/CVF International Conference on Computer Vision (CVPR)}, 2021.

\bibitem{ouyang2022cosmo}
X.~Ouyang, X.~Shuai, J.~Zhou, I.~W. Shi, Z.~Xie, G.~Xing, and J.~Huang, ``Cosmo: Contrastive fusion learning with small data for multimodal human activity recognition,'' in \emph{International Conference on Mobile Computing And Networking (MobiCom)}, 2022.

\bibitem{poklukar2022gmc}
P.~Poklukar, M.~Vasco, H.~Yin, F.~S. Melo, A.~Paiva, and D.~Kragic, ``Geometric multimodal contrastive representation learning,'' in \emph{International Conference on Machine Learning (ICML)}, 2022.

\bibitem{tonekaboni2021unsupervised}
S.~Tonekaboni, D.~Eytan, and A.~Goldenberg, ``Unsupervised representation learning for time series with temporal neighborhood coding,'' in \emph{International Conference on Learning Representations (ICLR)}, 2021.

\bibitem{yue2022ts2vec}
Z.~Yue, Y.~Wang, J.~Duan, T.~Yang, C.~Huang, Y.~Tong, and B.~Xu, ``Ts2vec: Towards universal representation of time series,'' in \emph{AAAI Conference on Artificial Intelligence (AAAI)}, 2022.

\bibitem{chen2020simclr}
T.~Chen, S.~Kornblith, M.~Norouzi, and G.~Hinton, ``A simple framework for contrastive learning of visual representations,'' in \emph{International Conference on Machine Learning (ICML)}, 2020.

\bibitem{tian2020contrastive}
Y.~Tian, D.~Krishnan, and P.~Isola, ``Contrastive multiview coding,'' in \emph{European Conference on Computer Vision (ECCV)}, 2020.

\bibitem{li2021faster}
J.~Li, M.~Jing, H.~Su, K.~Lu, L.~Zhu, and H.~T. Shen, ``Faster domain adaptation networks,'' \emph{IEEE Transactions on Knowledge and Data Engineering}, vol.~34, no.~12, pp. 5770--5783, 2021.

\bibitem{zhao2022memory}
Y.~Zhao, D.~Saxena, and J.~Cao, ``Memory-efficient domain incremental learning for internet of things,'' in \emph{Proceedings of the 20th ACM Conference on Embedded Networked Sensor Systems}, 2022, pp. 1175--1181.

\bibitem{zhang2023federated}
L.~Zhang, X.~Lei, Y.~Shi, H.~Huang, and C.~Chen, ``Federated learning for iot devices with domain generalization,'' \emph{IEEE Internet of Things Journal}, 2023.

\bibitem{huang2022incremental}
Y.~Huang, M.~Du, H.~Zheng, and X.~Feng, ``Incremental unsupervised adversarial domain adaptation for federated learning in iot networks,'' in \emph{2022 18th International Conference on Mobility, Sensing and Networking (MSN)}.\hskip 1em plus 0.5em minus 0.4em\relax IEEE, 2022, pp. 186--190.

\bibitem{achituve2021self}
I.~Achituve, H.~Maron, and G.~Chechik, ``Self-supervised learning for domain adaptation on point clouds,'' in \emph{Proceedings of the IEEE/CVF Winter Conference on Applications of Computer Vision}, 2021, pp. 123--133.

\bibitem{xu2019self}
J.~Xu, L.~Xiao, and A.~M. L{\'o}pez, ``Self-supervised domain adaptation for computer vision tasks,'' \emph{IEEE Access}, vol.~7, pp. 156\,694--156\,706, 2019.

\end{thebibliography}

\end{document}